\documentclass[australian,english]{article}
\usepackage[latin9]{inputenc}
\usepackage{babel}
\usepackage{url}
\usepackage{multirow}
\usepackage{amsmath}
\usepackage{amssymb}
\usepackage{graphicx}
\usepackage{xargs}[2008/03/08]
\usepackage[unicode=true,
 bookmarks=false,
 breaklinks=false,pdfborder={0 0 0},pdfborderstyle={},backref=section,colorlinks=false]
 {hyperref}

\makeatletter

\providecommand{\tabularnewline}{\\}

\usepackage{RobustML_iclr2021_conference} 
\usepackage{times}

\usepackage{url}

\title{Improved and Efficient Text Adversarial \\ Attacks using Target Information}

\author{Mahmoud Hossam, Trung Le, He Zhao, Viet Huynh \& Dinh Phung \\
Department of Data Science and AI, Faculty of Information Technology\\
Monash University\\
Clayton 3800, Victoria, Australia\\
\texttt{\{mhossam, trunglm, ethan.zhao, viet.huynh, dinh.phung\}@monash.edu}
}

\newcommand{\new}{\marginpar{NEW}}

\renewcommand{\cite}{\citep}
\renewcommand{\citet}{\citep}

\iclrfinalcopy 

\makeatother

\begin{document}
\maketitle
\begin{abstract}
There has been recently a growing interest in studying adversarial
examples on natural language models in the black-box setting. These
methods attack natural language classifiers by perturbing certain
important words until the classifier label is changed. In order to
find these important words, these methods rank all words by importance
by querying the target model word by word for each input sentence,
resulting in high query inefficiency. A new interesting approach was
introduced that addresses this problem through interpretable learning
to learn the word ranking instead of previous expensive search. The
main advantage of using this approach is that it achieves comparable
attack rates to the state-of-the-art methods, yet faster and with
fewer queries, where fewer queries are desirable to avoid suspicion
towards the attacking agent. Nonetheless, this approach sacrificed
the useful information that could be leveraged from the target classifier
for the sake of query efficiency. In this paper we study the effect
of leveraging the target model outputs and data on both attack rates
and average number of queries, and we show that both can be improved,
with a limited overhead of additional queries.  
\end{abstract}
\selectlanguage{australian}%
\newcommand{\sidenote}[1]{\marginpar{\small \emph{\color{Medium}#1}}}

\global\long\def\se{\hat{\text{se}}}%

\global\long\def\interior{\text{int}}%

\global\long\def\boundary{\text{bd}}%

\global\long\def\ML{\textsf{ML}}%

\global\long\def\MAP{\textsf{MAP}}%

\global\long\def\GML{\mathsf{GML}}%

\global\long\def\HMM{\mathsf{HMM}}%

\global\long\def\support{\text{supp}}%

\global\long\def\new{\text{*}}%

\global\long\def\stir{\text{Stirl}}%

\global\long\def\eA{\mathcal{A}}%

\global\long\def\eB{\mathcal{B}}%

\global\long\def\eC{\mathcal{C}}%

\global\long\def\eD{\mathcal{D}}%

\global\long\def\eE{\mathcal{E}}%

\global\long\def\eF{\mathcal{F}}%

\global\long\def\eG{\mathcal{G}}%

\global\long\def\eH{\mathcal{H}}%

\global\long\def\eI{\mathcal{I}}%

\global\long\def\eJ{\mathcal{J}}%

\global\long\def\eK{\mathcal{K}}%

\global\long\def\eL{\mathcal{L}}%

\global\long\def\eM{\mathcal{M}}%

\global\long\def\eN{\mathcal{N}}%

\global\long\def\eO{\mathcal{O}}%

\global\long\def\eP{\mathcal{P}}%

\global\long\def\eQ{\mathcal{Q}}%

\global\long\def\eR{\mathcal{R}}%

\global\long\def\eS{\mathcal{S}}%

\global\long\def\eT{\mathcal{T}}%

\global\long\def\eU{\mathcal{U}}%

\global\long\def\eV{\mathcal{V}}%

\global\long\def\eW{\mathcal{W}}%

\global\long\def\eX{\mathcal{X}}%

\global\long\def\eY{\mathcal{Y}}%

\global\long\def\eZ{\mathcal{Z}}%

\global\long\def\dist{d}%

\global\long\def\HX{\entro\left(X\right)}%
 
\global\long\def\entropyX{\HX}%

\global\long\def\HY{\entro\left(Y\right)}%
 
\global\long\def\entropyY{\HY}%

\global\long\def\HXY{\entro\left(X,Y\right)}%
 
\global\long\def\entropyXY{\HXY}%

\global\long\def\mutualXY{\mutual\left(X;Y\right)}%
 
\global\long\def\mutinfoXY{\mutualXY}%

\global\long\def\given{\mid}%

\global\long\def\gv{\given}%

\global\long\def\goto{\rightarrow}%

\global\long\def\asgoto{\stackrel{a.s.}{\longrightarrow}}%

\global\long\def\pgoto{\stackrel{p}{\longrightarrow}}%

\global\long\def\dgoto{\stackrel{d}{\longrightarrow}}%

\global\long\def\lik{\mathcal{L}}%

\global\long\def\logll{\mathit{l}}%

\global\long\def\vectorize#1{\mathbf{#1}}%

\global\long\def\vt#1{\mathbf{#1}}%

\global\long\def\gvt#1{\boldsymbol{#1}}%

\global\long\def\idp{\ \bot\negthickspace\negthickspace\bot\ }%
 
\global\long\def\cdp{\idp}%

\global\long\def\das{:=}%

\global\long\def\id{\mathbb{I}}%

\global\long\def\idarg#1#2{\id\left\{  #1,#2\right\}  }%

\global\long\def\iid{\stackrel{\text{iid}}{\sim}}%

\global\long\def\idpdraw{\stackrel{\text{indep}}{\sim}}%

\global\long\def\bzero{\vt 0}%

\global\long\def\bone{\mathbf{1}}%

\global\long\def\boldm{\boldsymbol{m}}%

\global\long\def\bff{\vt f}%

\global\long\def\bx{\boldsymbol{x}}%

\global\long\def\ba{\boldsymbol{a}}%

\global\long\def\bb{\boldsymbol{b}}%

\global\long\def\bc{\boldsymbol{c}}%

\global\long\def\bl{\boldsymbol{l}}%

\global\long\def\bu{\boldsymbol{u}}%

\global\long\def\bo{\boldsymbol{o}}%

\global\long\def\bh{\boldsymbol{h}}%

\global\long\def\bs{\boldsymbol{s}}%

\global\long\def\bz{\boldsymbol{z}}%

\global\long\def\xnew{y}%

\global\long\def\bxnew{\boldsymbol{y}}%

\global\long\def\bX{\boldsymbol{X}}%

\global\long\def\tbx{\tilde{\bx}}%

\global\long\def\by{\boldsymbol{y}}%

\global\long\def\bY{\boldsymbol{Y}}%

\global\long\def\bZ{\boldsymbol{Z}}%

\global\long\def\bU{\boldsymbol{U}}%

\global\long\def\bv{\boldsymbol{v}}%

\global\long\def\bn{\boldsymbol{n}}%

\global\long\def\bV{\boldsymbol{V}}%

\global\long\def\bI{\boldsymbol{I}}%

\global\long\def\bw{\vt w}%

\global\long\def\bW{\boldsymbol{W}}%

\global\long\def\balpha{\gvt{\alpha}}%

\global\long\def\bbeta{\gvt{\beta}}%

\global\long\def\bmu{\gvt{\mu}}%

\global\long\def\btheta{\boldsymbol{\theta}}%

\global\long\def\blambda{\boldsymbol{\lambda}}%

\global\long\def\bgamma{\boldsymbol{\gamma}}%

\global\long\def\bpsi{\boldsymbol{\psi}}%

\global\long\def\bphi{\boldsymbol{\phi}}%

\global\long\def\bpi{\boldsymbol{\pi}}%

\global\long\def\bomega{\boldsymbol{\omega}}%

\global\long\def\bepsilon{\boldsymbol{\epsilon}}%

\global\long\def\btau{\boldsymbol{\tau}}%

\global\long\def\bxi{\boldsymbol{\xi}}%

\global\long\def\va{\vectorize a}%

\global\long\def\vb{\vectorize b}%

\global\long\def\vc{\vectorize c}%

\global\long\def\vt{\vectorize t}%

\global\long\def\vx{\vectorize x}%

\global\long\def\vy{\vectorize y}%

\global\long\def\vz{\vectorize z}%

\global\long\def\realset{\mathbb{R}}%

\global\long\def\realn{\realset^{n}}%

\global\long\def\naturalset{\mathbb{N}}%

\global\long\def\natn{\mathbb{N}^{n}}%

\global\long\def\integerset{\mathbb{Z}}%

\global\long\def\intn{\integerset^{n}}%

\global\long\def\rational{\mathbb{Q}}%

\global\long\def\rationaln{\rational^{n}}%

\global\long\def\complexset{\mathbb{C}}%

\global\long\def\comp{\complexset}%

\global\long\def\compl#1{#1^{\text{c}}}%

\global\long\def\and{\cap}%

\global\long\def\compn{\comp^{n}}%

\global\long\def\comb#1#2{\left({#1\atop #2}\right) }%

\global\long\def\nchoosek#1#2{\left({#1\atop #2}\right)}%

\global\long\def\param{\vt w}%

\global\long\def\Param{\Theta}%

\global\long\def\meanparam{\gvt{\mu}}%

\global\long\def\Meanparam{\mathcal{M}}%

\global\long\def\meanmap{\mathbf{m}}%

\global\long\def\logpart{A}%

\global\long\def\simplex{\Delta}%

\global\long\def\simplexn{\simplex^{n}}%

\global\long\def\dirproc{\text{DP}}%

\global\long\def\ggproc{\text{GG}}%

\global\long\def\DP{\text{DP}}%

\global\long\def\ndp{\text{nDP}}%

\global\long\def\hdp{\text{HDP}}%

\global\long\def\gempdf{\text{GEM}}%

\global\long\def\rfs{\text{RFS}}%

\global\long\def\bernrfs{\text{BernoulliRFS}}%

\global\long\def\poissrfs{\text{PoissonRFS}}%

\global\long\def\iidrfs{$\text{IidRFS}$}%

\global\long\def\grad{\gradient}%
 
\global\long\def\gradient{\nabla}%

\global\long\def\partdev#1#2{\partialdev{#1}{#2}}%
 
\global\long\def\partialdev#1#2{\frac{\partial#1}{\partial#2}}%

\global\long\def\partddevvect#1#2{\partialdevdevvect{#1}{#2}}%
 
\global\long\def\partialdevdevvect#1#2{\frac{\partial^{2}#1}{\partial#2\partial#2^{\top}}}%

\global\long\def\partddev#1#2{\partialdevdev{#1}{#2}}%
 
\global\long\def\partialdevdev#1#2{\frac{\partial^{2}#1}{\partial#2^{2}}}%

\global\long\def\dev#1{\mathrm{d}#1}%

\global\long\def\hessian{\text{Hess}}%

\global\long\def\closure{\text{cl}}%

\global\long\def\realpart{\text{Re }}%

\global\long\def\impart{\text{Im }}%

\global\long\def\probP{\mathbb{P}}%

\global\long\def\cpr#1#2{\Pr\left(#1\ |\ #2\right)}%

\global\long\def\var{\text{Var}}%

\global\long\def\Var#1{\text{Var}\left[#1\right]}%

\global\long\def\cov{\text{Cov}}%

\global\long\def\Cov#1{\cov\left[ #1 \right]}%

\global\long\def\COV#1#2{\underset{#2}{\cov}\left[ #1 \right]}%

\global\long\def\corr{\text{Corr}}%

\global\long\def\sst{\text{T}}%

\global\long\def\SST{\sst}%

\global\long\def\ess{\mathbb{E}}%

\global\long\def\Ess#1{\mathbb{E}\left[#1\right]}%

\newcommandx\ESS[2][usedefault, addprefix=\global, 1=]{\underset{#2}{\mathbb{E}}\left[#1\right]}%

\global\long\def\fisher{\mathcal{F}}%

\global\long\def\bfield{\mathcal{B}}%
 
\global\long\def\borel{\mathcal{B}}%

\global\long\def\law{\text{law}}%

\global\long\def\bernpdf{\text{Bern}}%

\global\long\def\bernoullipdf{\text{Bernoulli}}%

\global\long\def\binpdf{\text{Bin}}%

\global\long\def\binomialpdf{\text{Binomial}}%

\global\long\def\betapdf{\text{Beta}}%

\global\long\def\dirpdf{\text{Dir}}%

\global\long\def\dirichletpdf{\text{Dirichlet}}%

\global\long\def\gammapdf{\text{Gamma}}%

\global\long\def\gaussden#1#2{\mathcal{N}\left(#1, #2 \right) }%

\global\long\def\gauss{\mathcal{N}}%

\global\long\def\gausspdf#1#2#3{\mathcal{N}\left( #1 \lcabra{#2, #3}\right) }%

\global\long\def\multpdf{\text{Mult}}%

\global\long\def\multinomialpdf{\text{Multinomial}}%

\global\long\def\catpdf{\text{Cat}}%

\global\long\def\categoricalpdf{\text{Categorical}}%

\global\long\def\poisspdf{\text{Pois}}%

\global\long\def\poissonpdf{\text{Poisson}}%

\global\long\def\pgpdf{\text{PG}}%

\global\long\def\wshpdf{\text{Wish}}%

\global\long\def\iwshpdf{\text{InvWish}}%

\global\long\def\nwpdf{\text{NW}}%

\global\long\def\niwpdf{\text{NIW}}%

\global\long\def\studentpdf{\text{Student}}%

\global\long\def\unipdf#1#2{\mathcal{U}\left(#1,#2\right)}%

\global\long\def\uniformpdf#1#2{\text{Uniform}\left(#1,#2\right)}%

\global\long\def\transp#1{\transpose{#1}}%
 
\global\long\def\transpose#1{#1^{\mathsf{T}}}%

\global\long\def\mgt{\succ}%

\global\long\def\mge{\succeq}%

\global\long\def\idenmat{\mathbf{I}}%

\global\long\def\trace{\mathrm{tr}}%

\global\long\def\argmax#1{\underset{_{#1}}{\text{argmax}\;} }%

\global\long\def\argmin#1{\underset{_{#1}}{\text{argmin}\ } }%

\global\long\def\diag{\text{diag}}%

\global\long\def\norm{}%

\global\long\def\comp{\mathrm{comp}}%

\global\long\def\proj{\mathrm{proj}}%

\global\long\def\spn{\text{span}}%

\global\long\def\vtspace{\mathcal{V}}%

\global\long\def\field{\mathcal{F}}%
 
\global\long\def\ffield{\mathcal{F}}%

\global\long\def\inner#1#2{\left\langle #1,#2\right\rangle }%
 
\global\long\def\iprod#1#2{\inner{#1}{#2}}%

\global\long\def\dprod#1#2{#1 \cdot#2}%

\global\long\def\norm#1{\left\Vert #1\right\Vert }%

\global\long\def\entro{\mathbb{H}}%

\global\long\def\entropy{\mathbb{H}}%

\global\long\def\Entro#1{\entro\left[#1\right]}%

\global\long\def\Entropy#1{\Entro{#1}}%

\global\long\def\mutinfo{\mathbb{I}}%

\global\long\def\relH{\mathit{D}}%

\global\long\def\reldiv#1#2{\relH\left(#1||#2\right)}%

\global\long\def\KL{KL}%

\global\long\def\JS{JS}%

\global\long\def\fdiv#1#2{D_{f}\left(#1\parallel#2\right)}%
 
\global\long\def\fdivergence#1#2{D_{f}\left(#1\ \parallel\ #2\right)}%

\global\long\def\KLdiv#1#2{\KL\left(#1\parallel#2\right)}%
 
\global\long\def\KLdivergence#1#2{\KL\left(#1\ \parallel\ #2\right)}%

\global\long\def\JSdiv#1#2{\JS\left(#1\parallel#2\right)}%
 
\global\long\def\JSdivergence#1#2{\JS\left(#1\ \parallel\ #2\right)}%

\global\long\def\crossH{\mathcal{C}}%
 
\global\long\def\crossentropy{\mathcal{C}}%

\global\long\def\crossHxy#1#2{\crossentropy\left(#1\parallel#2\right)}%

\global\long\def\breg{\text{BD}}%

\global\long\def\stein{\mathbb{S}}%

\global\long\def\lcabra#1{\left|#1\right.}%

\global\long\def\lbra#1{\lcabra{#1}}%

\global\long\def\rcabra#1{\left.#1\right|}%

\global\long\def\rbra#1{\rcabra{#1}}%
\selectlanguage{english}%

\section{Introduction}

Most of the existing black-box attack methods on natural language
models directly query the target model to rank sentence words for
replacement. This procedure is expensive in terms of number of queries
needed to the target model, and the number of queries increases with
increasing input length. In addition, the increased number of queries
to the target classifier is not desirable in black-box settings, where
it can raise the suspicion towards the attacking agent. Recently,
several methods were introduced to improve the query efficiency, including
\cite{garg2020bae,li2020bertattack} and \textit{Explain2Attack} \cite{hossam2021explain2attack},
where an interpretable model was trained to learn word importance
ranking for synonym replacement. 

Unlike the other methods, Explain2Attack addresses the word ranking
query inefficiency, which is closely related to the target model,
and causes a hurdle into efficiently crafting the adversarial example.
In Explain2Attack model, a \textit{selector} network $\mathcal{E}$
is trained to learn the important words by approximating the performance
of a \textit{substitute classifier} $F_{b}$ . Under this architecture,
Explain2Attack was able to achieve comparable attack rates to the
state-of-the-art method \cite{DBLP:journals/corr/abs-1907-11932},
while reducing the number of queries. 

However, the word ranking step in Explain2Attack was trained on the
substitute dataset and classifier, which are both different from the
target classifier and the target dataset. The substitute training
was designed in this way in order to limit the access to the target
model through extensive queries, and to its training dataset $D_{t}^{\textrm{train}}$
. The key intuition is that as long as the chosen dataset for the
substitute training dataset $D_{b}$ comes from a similar domain to
the target training dataset, then the learned word ranking model should
perform well enough on test sentences from the target dataset domain. 

Although this setting leads to competitive attack rates and reduced
queries, it is still a restriction on the substitute training procedure.
Additionally, in some scenarios, the access to target model data might
be available. Thus, by relaxing restrictions on the access to possible
target model information (both outputs and data domain), we might
be able to achieve better attack rates with a marginal overhead of
additional queries used for the substitute training. 

In this paper, we propose two modifications to this interpretable
approach to better train the selector network while benefiting from
available target information. Specifically, we do not use the substitute
dataset labels used for selector training as in \cite{hossam2021explain2attack}.
Instead, we replace them with output predictions from the target classifier
under two settings: the target model is given the substitute dataset
sentences $\mathcal{X}_{b}\in D_{b}$, or the target model is given
the target test set $\mathbb{\mathcal{X}}_{t}^{\textrm{test}}\in D_{t}^{\textrm{train}}$.

The advantage are twofold: \textit{i)} \textit{we relax the requirement
of finding a suitable substitute data domain to learn the word ranking},
and \textit{ii) we achieve both improved attack rates and reduced
average number of queries, with limited effect on the number of overhead
queries}. We describe in details the proposed modifications and
their intuitions, and provide experimental results for different classification
models and datasets. 

\section{Background }

Here we revise the original setting of Explain2Attack \cite{hossam2021explain2attack}:
let a target model $F_{\textrm{target}}$ be trained on some target
training dataset $D_{t}^{\textrm{train}}=\left\{ \mathbb{\mathcal{X}}_{t}^{\textrm{train}},\mathcal{Y}_{t}^{\textrm{train}}\right\} $
and testing set $D_{t}^{\textrm{test}}=\left\{ \mathbb{\mathcal{X}}_{t}^{\textrm{test}},\mathcal{Y}_{t}^{\textrm{test}}\right\} $.
A substitute model, $\textrm{SUB}$, is then trained to learn the
word importance scores. $\textrm{SUB}$ is trained using a dataset
that is close enough to the target model dataset called the substitute
dataset, $D_{b}=\left\{ \mathcal{X}_{b},\mathcal{Y}_{b}\right\} $.
The substitute model itself contains two sub-networks called the \textit{substitute
classifier} $F_{b}$ and the \textit{selector} network $\mathcal{E}_{\theta}$.
After training, performing inference on the selector network $\mathcal{E}_{\theta}(X)$
using input sentences yields the desired word importance scores for
these sentences. During the substitute training procedure, the substitute
classifier $F_{b}$ is trained to correctly predict the label $Y\in\mathcal{Y}_{b}$
from an input sentence $X_{s}\sim\mathcal{E}_{\theta}(X),$ where
$X\in\mathcal{X}_{b}$, and $X_{s}$ is the most important selected
$k$ words for some input $X$. After substitute training is finished,
$F_{b}(.)$ can be discarded, since we are only interested in the
selector $\mathcal{E}(X)$.

\section{Method}

We change the original setting of Explain2Attack to incorporate the
target classifier predictions outputs in the selector network training.
Specifically, we consider two black--box settings: first, the substitute
networks are trained without access to the target test set, and second,
they are trained with access to the target test set. 

In details, we want to replace the labels $\mathcal{Y}_{b}$ that
come from the substitute dataset $D_{b}$ with output labels or probabilities
from the target classifier $F_{\textrm{target}}$. This way, the training
of $F_{b}$, and most importantly, the selector $\mathcal{E}_{\theta}$,
will be trained such that it learns or imitates the behaviour of the
target classifier. This requires querying the target model $F_{\textrm{target}}\left(\widetilde{X}_{\textrm{}}\right)$
with some input $\widetilde{X}$ to return target classifier predictions.
The choice of the inputs to the target model $\widetilde{X}$ allows
to different settings: \textit{i)} We either choose $\widetilde{X}_{\textrm{}}$
to come from the substitute dataset $\mathcal{X}_{b}$, or \textit{ii)}
from the target test-set sentences $\mathbb{\mathcal{X}}_{t}^{\textrm{test}}$.
Below we describe both settings in details and the possible use cases
for both.

\subsection{\label{subsec:chp6_Substitute-training-without}Substitute training
without access to the target test set}

\begin{figure*}
\begin{centering}
\includegraphics[viewport=300bp 0bp 800bp 540bp,clip,scale=0.35]{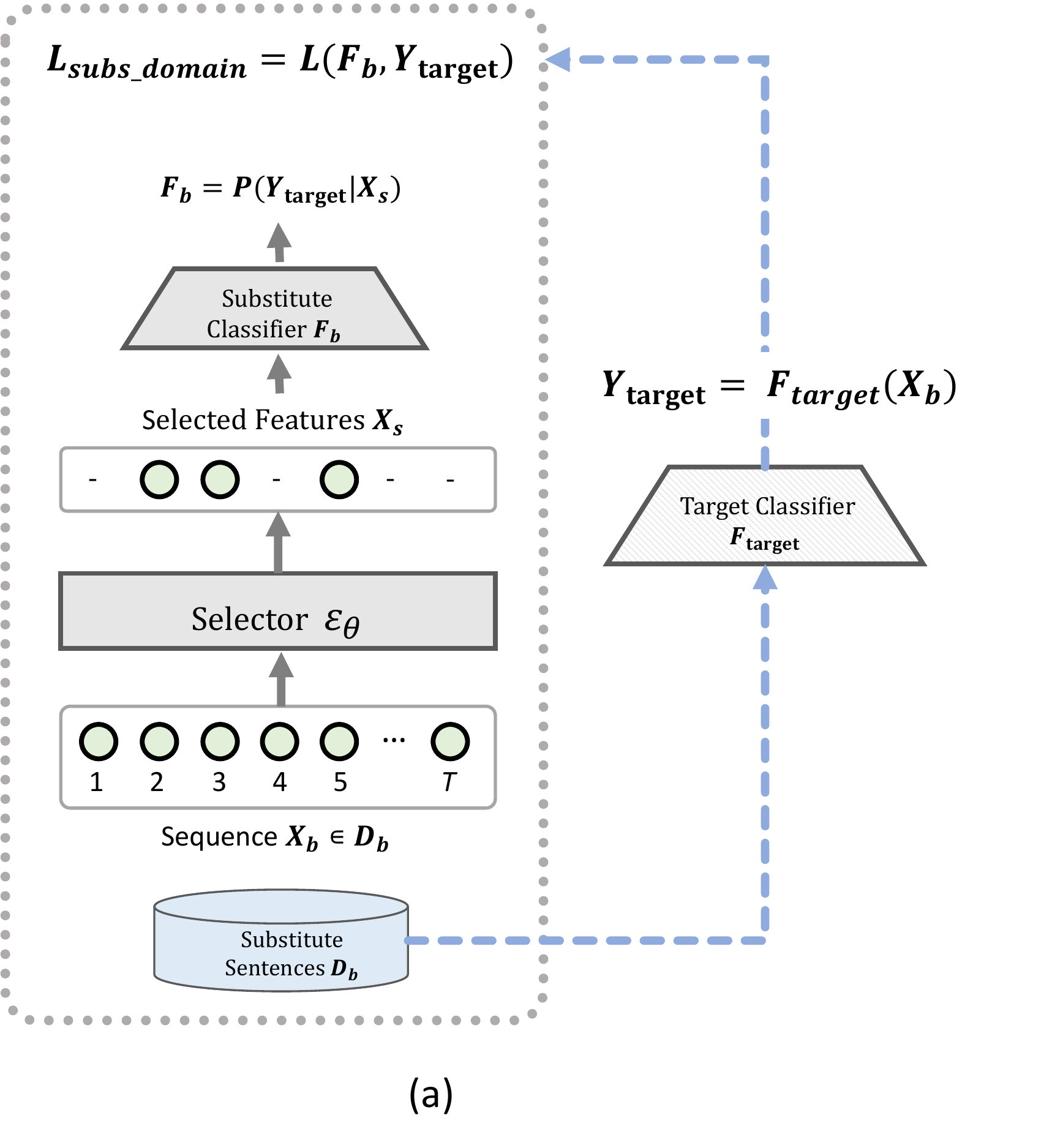}\includegraphics[viewport=300bp 0bp 800bp 540bp,clip,scale=0.35]{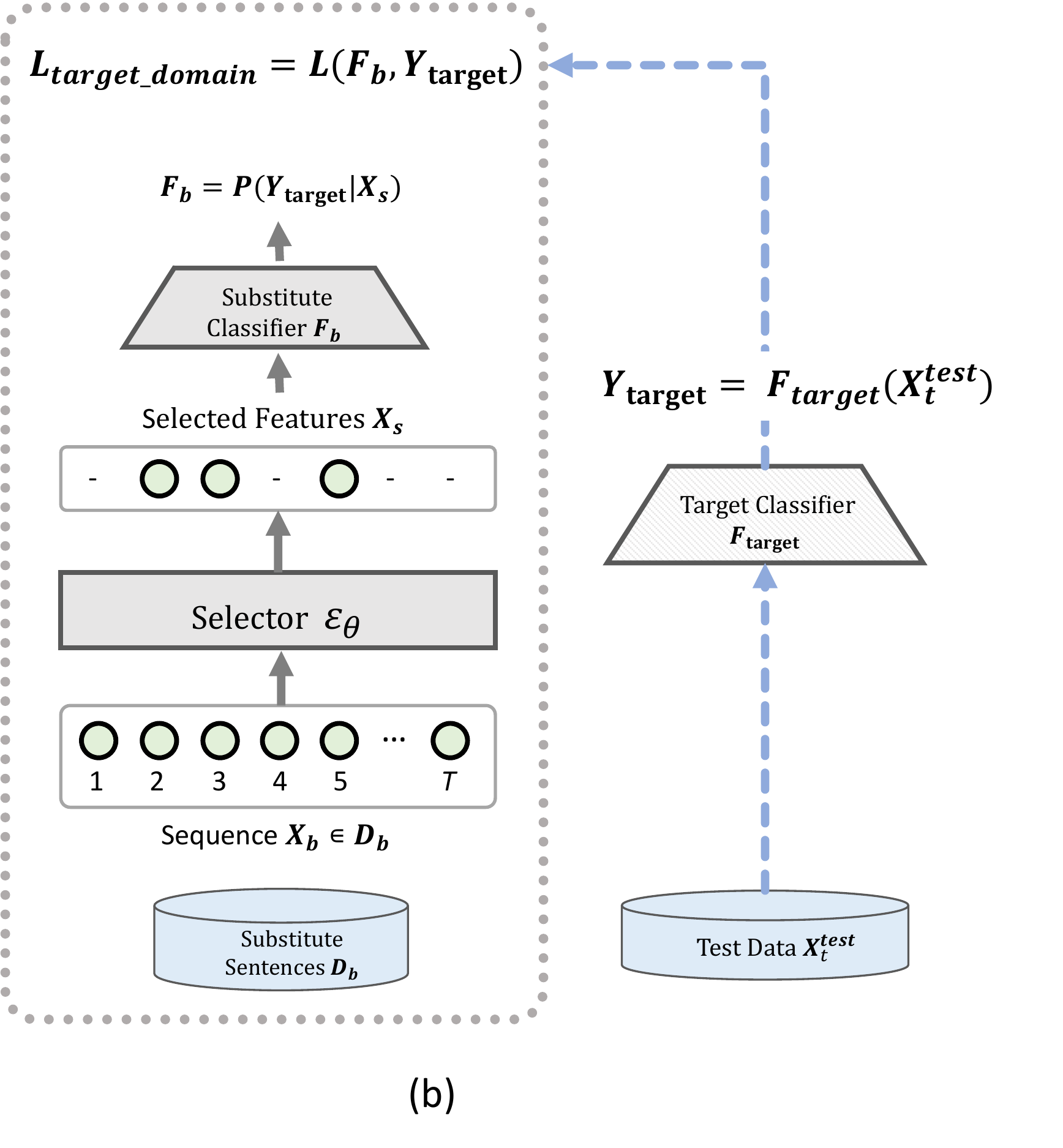}
\par\end{centering}
\caption{\label{fig:chp6_framework_1}Overview of modified Explain2Attack training
with target model information via a) the substitute sentences, or
b) the target test set.}
\end{figure*}

In this setting, we use the target model's predictions to come from
substitute sentences input $X_{b}\in\mathcal{X}_{b}$ instead of using
the substitute dataset labels $\mathcal{Y}_{b}$. We use the prediction
probabilities from $F_{\textrm{target}}$ for the substitute training,
where the loss function for training both $F_{b}$ and $\mathcal{E}_{\theta}$
becomes:

{\small{}
\begin{equation}
\mathcal{L}_{\textrm{subs\_domain}}\left(F_{b},Y_{\textrm{target}}\right)=\ESS[\left\Vert F_{b_{\phi}}(\mathcal{E}_{\theta}(X_{b}))-F_{\textrm{target}}(X_{b})\right\Vert _{2}^{2}]{X_{b}\in D_{b}},\label{eq:chp6_method_1_mse}
\end{equation}
}where $D_{b}$ is the substitute dataset, $F_{b}\left(\mathcal{E}_{\theta}\left(\cdot\right)\right)$
is the substitute classifier output, and $Y_{\textrm{target}}$ is
the output of the target classifier $F_{\textrm{target}}$, both containing
the probabilities for all available classes. This loss is the mean
square error (MSE) between the substitute classifier and the target
classifier outputs. Unlike other losses like the KL-divergence, the
information about the other classes in the soft label outputs (all
classes probabilities) are included. Therefore the purpose of using
the MSE loss is to train the substitute classifier such that its real-valued
output predictions (all of the classes probabilities) are optimized
to resemble the target classifier prediction outputs. Therefore, the
selector network is trained to help the substitute classifier resemble
the target classifier behavior. Fig. (\ref{fig:chp6_framework_1}a)
shows the substitute training procedure under this setting. This
setting is suitable when the target test set in not known for the
attacker, which is the common scenario of black-box attacks on pretrained
models.

\subsection{\label{subsec:chp6_Substitute-training-with}Substitute training
with access to the target test set}

In this setting, we use the target model's outputs to come from a
part of the target test set $X_{t}^{\textrm{test}}\in\mathbb{\mathcal{X}}_{t}^{\textrm{test}}$.
The rationale behind this setting is that we consider the cases when
the attacker happens to have access to the target dataset, or that
the user of the target model gets to choose the training dataset,
yet does not have access to the final target model parameters. Similar
to the previous setting, we use the probability outputs from $F_{\textrm{target}}$
for the substitute training, where the loss function becomes:

{\small{}
\begin{equation}
\mathcal{L}_{\textrm{target\_domain}}\left(F_{b},Y_{\textrm{target}}\right)=\ESS[\left\Vert F_{b_{\phi}}(\mathcal{E}_{\theta}(X_{b}))-F_{\textrm{target}}\left(X_{t}^{\textrm{test}}\right)\right\Vert _{2}^{2}]{X_{b}\in D_{b},\,X_{t}^{\textrm{test}}\in\mathbb{\mathcal{X}}_{t}^{\textrm{test}}}.\label{eq:chp6_method_2_mse}
\end{equation}
}{\small\par}

It is important to mention that in Eq.(\ref{eq:chp6_method_2_mse})
$X_{t}^{\textrm{test}}$ could be used instead of $X_{b}$ in principle,
however, we show that even using data other than the target test still
improves the results. Fig. (\ref{fig:chp6_framework_1}b) shows the
substitute training procedure under this setting. 

\section{Experiments and Discussion}

We employ target classifier predictions under these two settings on
sentiment classification task using WordCNN and WordLSTM target classifiers.
For all of the experiments, we used the same datasets, substitute
datasets, and target classifier parameters and hyperparameters that
were used in \cite{hossam2021explain2attack}. 

To evaluate the performance of the two proposed methods described
above, we report the adversarial accuracy ($\textrm{Adv\_Acc}$),
the average number of queries ($\textrm{Adv\_Queries}$), and compare
with the original Explain2Attack reported in \cite{hossam2021explain2attack}.
In detail, in tables \ref{tab:ch6_subs_Accuracy_Queries} and \ref{tab:ch6_target_Accuracy_Queries}
we report the results for incorporating the target model predictions
for training Explain2Attack selector without access to the target
test set and with access to it, respectively. For both $\textrm{Adv\_Acc}$
and $\textrm{Adv\_Queries}$, the lower number is the better, indicated
by the $\downarrow$ symbol.

For the first method, we use the exact setup as in \cite{hossam2021explain2attack}
for the experiments and the datasets, except that the substitute labels
$\mathcal{Y}_{b}$ were replaced with the target classifier predictions
from the substitute sentences, and Eq. (\ref{eq:chp6_method_1_mse})
was used for training the selector. For the second method, $\mathcal{Y}_{b}$
were replaced with the target classifier predictions from the target
test set, and Eq. (\ref{eq:chp6_method_2_mse}) was used to train
the selector. For the set of experiments in Table \ref{tab:ch6_target_Accuracy_Queries},
only a small portion of the test set was used during the selector
training, where 5000 out of 25,000 sentences were used. 

For the set of experiments in Table \ref{tab:ch6_subs_Accuracy_Queries},
the first method was used to attack the target classifier. We can
see that in three out of four experiments, that the attack rate was
improved compared to the original baseline. This demonstrates the
added benefit of incorporating the target model predictions in the
training process. Moreover, we find that the average number of queries
needed was also improved in these three cases.

\begin{table}[h]
\caption{\label{tab:ch6_subs_Accuracy_Queries}Performance metrics for Explain2Attack
(E2A) with the selector trained on target model predictions given
substitute dataset sentences.}

\medskip{}

\begin{centering}
\resizebox{1.00\textwidth}{!}{
\begin{tabular}{cc|cc|cc}
\hline 
\multicolumn{2}{c}{{\small{}Classifier}} & \multicolumn{2}{c}{{\small{}WordCNN}} & \multicolumn{2}{c}{{\small{}WordLSTM}}\tabularnewline
\hline 
\multicolumn{2}{c|}{{\small{}Target Model}} & {\small{}IMDB} & {\small{}Amazon MR} & {\small{}IMDB} & {\small{}Amazon MR}\tabularnewline
\hline 
 & {\small{}$\textrm{Clean\_Acc}$.} & 87.32 & 90.16 & 88.78 & 91.44\tabularnewline
\hline 
\multirow{3}{*}{{\small{}Adv\_Acc. $\downarrow$}} & \textit{\footnotesize{}(Substitute Data)} & \multicolumn{1}{c|}{\textit{\footnotesize{}(Yelp)}} & \multicolumn{1}{c|}{\textit{\footnotesize{}(IMDB)}} & \multicolumn{1}{c|}{\textit{\footnotesize{}(Amazon MR)}} & \textit{\footnotesize{}(IMDB)}\tabularnewline
 & E2A{\small{} } & 0.59 & 4.12 & \textbf{0.05} & 2.51\tabularnewline
 & E2A{\small{}-$F(X_{b})$ (ours)} & \textbf{0.57} & \textbf{4.11} & 0.08 & \textbf{2.37}\tabularnewline
\cline{2-6} \cline{3-6} \cline{4-6} \cline{5-6} \cline{6-6} 
\multirow{2}{*}{{\small{}$\textrm{Avg\_Queries}$ $\downarrow$}} & E2A & 402.5 & 351.7 & \textbf{440.2} & 368.7\tabularnewline
 & E2A{\small{}-$F(X_{b})$ (ours)} & \textbf{345.6} & \textbf{337.8} & 652.7 & \textbf{336.0}\tabularnewline
\end{tabular}}\caption{\label{tab:ch6_target_Accuracy_Queries}Performance metrics for Explain2Attack
(E2A) with the selector trained on target predictions given target
test set sentences (5K out of 25K target test sentences used).}
\par\end{centering}
\begin{raggedright}
\medskip{}
\par\end{raggedright}
\centering{}\resizebox{1.00\textwidth}{!}{
\begin{tabular}{cc|cc|cc}
\hline 
\multicolumn{2}{c}{{\small{}Classifier}} & \multicolumn{2}{c}{{\small{}WordCNN}} & \multicolumn{2}{c}{{\small{}WordLSTM}}\tabularnewline
\hline 
\multicolumn{2}{c|}{{\small{}Target Model}} & {\small{}IMDB} & {\small{}Amazon MR} & {\small{}IMDB} & {\small{}Amazon MR}\tabularnewline
\hline 
 & {\small{}$\textrm{Clean\_Acc}$.} & 87.32 & 90.16 & 88.78 & 91.44\tabularnewline
\hline 
\multirow{2}{*}{{\small{}Adv\_Acc. $\downarrow$}} & E2A & 0.59 & 4.12 & 0.05 & 2.51\tabularnewline
 & {\small{}E2A-$F\left(X_{t}^{\textrm{test}}\right)$ (ours)} & \textbf{0.56} & \textbf{4.05} & 0.05 & \textbf{2.34}\tabularnewline
\cline{2-6} \cline{3-6} \cline{4-6} \cline{5-6} \cline{6-6} 
\multirow{2}{*}{{\small{}$\textrm{Avg\_Queries}$ $\downarrow$}} & E2A & 402.5 & 351.7 & \textbf{440.2} & 368.7\tabularnewline
 & {\small{}E2A-$F\left(X_{t}^{\textrm{test}}\right)$ (ours)} & \textbf{382.6} & 352.2 & 444.4 & \textbf{357.2}\tabularnewline
\end{tabular}}
\end{table}

The improvement of the average number of queries is of particular
interest in our setting, as it relates to the quality of word importance
ranking learned during substitute training. Specifically, the baselines
TextFooler \cite{DBLP:journals/corr/abs-1907-11932} and the original
Explain2Attack both perform the following procedure for generating
an adversarial example: they perturb word by word, in order, from
the most to least important words. In each of these perturbations,
the target classifier is queried to check if its prediction label
was changed. Therefore, the correctness of word ranking plays a key
role in the total number of words that need to be perturbed, and consequently,
the total number of queries. This means that the more important the
selected words are to the target classifier, the fewer total words
will be needed to be perturbed in order to change the final classification
label. Thus, fewer queries will be needed. By looking on the results
in tables \ref{tab:ch6_subs_Accuracy_Queries} and \ref{tab:ch6_target_Accuracy_Queries},
we can see that for most of the experiments, there is a consistent
improvement in the number of queries. This observation suggests that
the employing target model predictions encouraged the selector to
learn more accurate word ranking according to its importance to the
target classifier. Although there is some overhead of queries involved
in training the selector in the first place, the later reduction in
average queries needed per single attack might be of more value with
the increased number of attacks.

In addition, we find that the overhead needed for substitute training
queries can be reduced in the second method, where the target test
set is used for output predictions from the target classifier. As
we see in Table \ref{tab:ch6_target_Accuracy_Queries}, the selector
was trained only on 5000 test set sentences, out of the total 25,000
sentences in the test set. This also suggests that having access to
the target test set might have an additional benefit on both attack
rates and the number of queries. We included a comprehensive ablation
study for different portion sizes used form the target dataset in
Appendix \ref{subsec:Appx_Target-Data-Size}.

\section{Conclusion}

In this paper, we studied the effect of incorporating the target model
domain data and outputs on attack rates and query efficiency in state-of-the-art
black--box text attacks. By leveraging target model output predictions
and test data, we presented two methods to improve the training process
of the recent interpretable learning approach Explain2Attack. We
use these target model outputs during the selector network training
instead of the substitute labels used in the original setting. The
intuition is to encourage the selector network to better learn the
behaviour of the target classifier through its outputs, thus learning
more accurate word importance ranking. The experiments show that on
most of the selected datasets, there was an improvement in both attack
rates and the average number of queries per attack. This study highlights
the benefit of carefully incorporating both target model data and
outputs in black-box attacks, while keeping a reduced average number
of queries per adversarial example. In the future, we plan to further
leverage the target classifier information by augmenting the substitute
training set with the intermediate adversarial candidates and their
target classifier outputs. \textbf{}

\subsubsection*{Acknowledgments}

This work was partially supported by the Australian Defence Science
and Technology (DST) Group under the Next Generation Technology Fund
(NTGF) scheme.

\textbf{\bibliography{references}
 \bibliographystyle{iclr2021_conference}}

\appendix

\section{Appendix}

\subsection{Related work}

Many methods were recently developed for black-box adversarial text
attacks \cite{kuleshov2018adversarial,yang2019greedy,gao2018black,ren-etal-2019-generating,DBLP:journals/corr/abs-1907-11932,garg2020bae,li2020bertattack}.
The main challenges for natural language adversarial attacks are the
discrete nature of inputs, where defining meaningful perturbations
is not straight forward, and the search space and complexity for finding
the important words.

In details, several methods \cite{DBLP:journals/corr/abs-1907-11932,ren-etal-2019-generating,gao2018black}
have been developed that share similar general framework, where the
attack starts by selecting the most important words/tokens to replace
from a candidate sentence, followed by searching for some word replacement
that can flip the classification label of the target model. However,
some methods followed the heuristic optimisation approach, for example,
\cite{alzantot-etal-2018-generating} used a genetic algorithm to
find the best sentence perturbation that fools the classifier. 

Most of the aforementioned black-box methods use the word selection/replacement
strategy. For instance, PWWS \cite{ren-etal-2019-generating} proposes
computing a word saliency score using output probabilities of the
target model, while \cite{gao2018black} computes sequential importance
score based on forward and backward RNN probabilities at the current
word position in the sentence. TextFooler \cite{DBLP:journals/corr/abs-1907-11932}
is a recent strong baseline for text attacks, where the method uses
a modified procedure for word ranking that increases the ranking in
label disagreement case. BERT-Attack and BAE-Attack \cite{li2020bertattack,garg2020bae}
both improve on TextFooler synonym replacement by using a pretrained
language model to generate suitable substitute words based on the
surrounding context. This achieved higher attack rates and the number
of queries is reduced. 

The recent introduced approach Explain2Attack \cite{hossam2021explain2attack}
differs from previous work in solving the word ranking problem. Unlike
other methods, instead of depending on the target model for word importance
ranking, word importance scores are \textit{learned}. The main differences
of this approach compared existing ones are: i) word ranking complexity
at inference time is of constant order in terms of input sentence
length, therefore, it significantly reduces the number of queries
needed to rank the words, as well as the running-time. ii) unlike
existing methods, this approach is scalable with increased sentence
lengths, since computing the scores is not dependent on word by word
query of the target model. Moreover, this general approach can benefit
from further query reduction in the synonym replacement phase by incorporating
the pretrained language model technique in methods like BERT-Attack
and BAE-Attack.

\subsection{\label{subsec:Appx_Target-Data-Size}Target Data Size Ablation Study}

\begin{figure*}
\begin{centering}
\includegraphics[viewport=0bp 50bp 710bp 540bp,clip,width=0.95\textwidth]{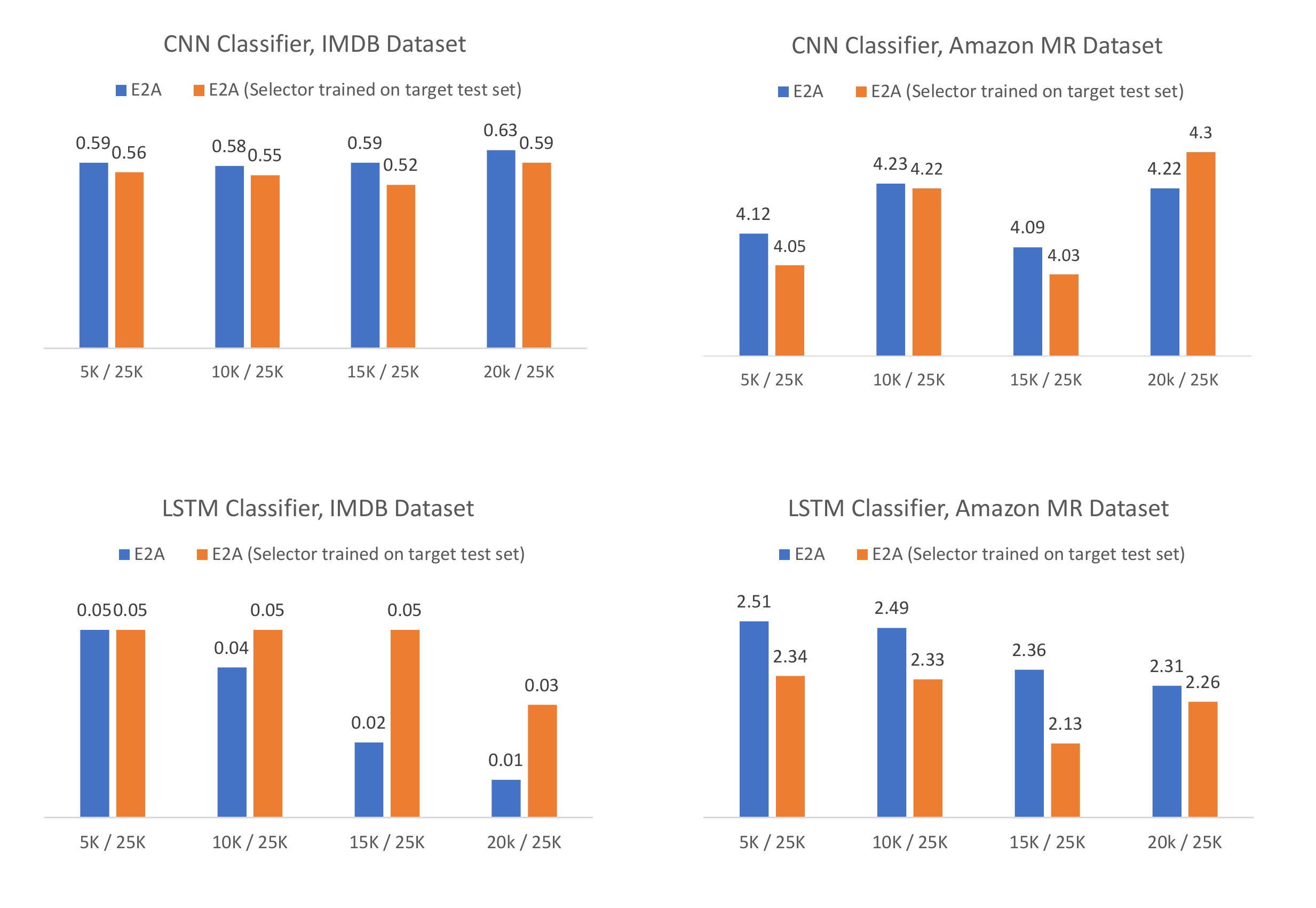}
\par\end{centering}
\caption[Adversarial Accuracies for Explain2Attack with the selector trained
on a portion of the target test set.]{\label{fig:chp6_target_accur_details}{\footnotesize{}Adversarial
Accuracies for Explain2Attack with the selector trained on a portion
of the target test set. Each category represents the portion of the
data used out of 25K samples. All combinations of target models and
datasets are reported (}\textit{\footnotesize{}lower is better}{\footnotesize{}
$\downarrow$).}}
\end{figure*}

\begin{figure*}
\begin{centering}
\includegraphics[viewport=0bp 50bp 760bp 540bp,clip,width=0.95\textwidth]{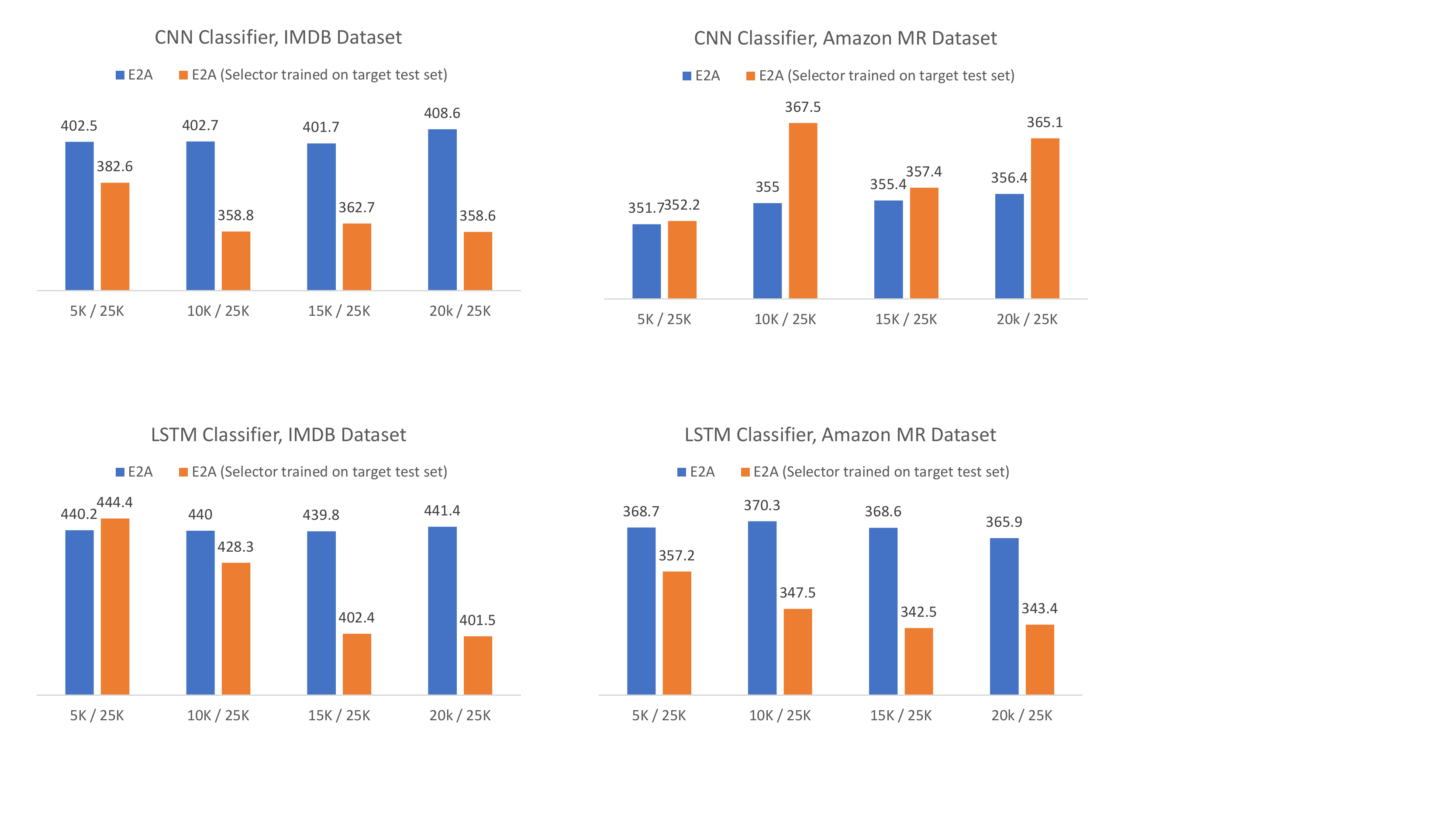}
\par\end{centering}
\caption[Average Queries for Explain2Attack with the selector trained on a
portion of the target test set.]{\label{fig:chp6_target_queries_details}{\footnotesize{}Average Queries
for Explain2Attack with the selector trained on a portion of the target
test set. Each category represents the portion of the data used out
of 25K samples. All combinations of target models and datasets are
reported (}\textit{\footnotesize{}lower is better}{\footnotesize{}
$\downarrow$).}}
\end{figure*}

We study the effect of the target set portion size used for the selector
training, using 5000, 10,000, 15,000, and 20,000 samples out of the
whole test set size of 25,000 samples. We perform the same set of
experiments in Table \ref{tab:ch6_target_Accuracy_Queries} on these
portions sizes and report the adversarial accuracies and average queries
for all combinations of target models and datasets in Figures (\ref{fig:chp6_target_accur_details})
and (\ref{fig:chp6_target_queries_details}) compared to the original
Expalin2Attack.

For every target model and dataset combination, we can see that there
is an improvement either in the adversarial accuracy (lower accuracy),
or in the average number of queries (fewer queries), or both. Notably,
we find that the best improvements happen mostly happen when the number
of used samples is less than or equal to 15,000. This suggests either
that there is a limit to the number of target test set samples that
can be useful for the selector training, or that this behavior is
just a special case on these selected datasets, and there might be
a more general consistent behavior under other datasets with different
sizes.

Similarly, we believe that the number of substitute dataset samples
used in the first method for training the selector might have an impact
on both attack rates and the number of queries. We look further to
investigate the impact of both target and substitute dataset sizes
on the overall performance by choosing more datasets with different
sizes, yet we leave this comprehensive study for future work.

\subsection{Datasets}

Here we briefly describe the datasets used in our experiments and
their statistics:
\begin{itemize}
\item \textbf{IMDB}: Movie reviews for sentiment classification \cite{maas2011learning,pang2005seeing}.
The reviews have binary labels, either positive or negative.
\item \textbf{Amazon MR}: Amazon polarity (binary) user reviews on movies,
extracted from the larger Amazon reviews polarity dataset \footnote{\url{https://www.kaggle.com/bittlingmayer/amazonreviews}}.
\item \textbf{Yelp Polarity Reviews}: Sentiment classification on positive
and negative businesses reviews \cite{zhang2015character}. We mainly
use this dataset as a substitute dataset when attacking other models.
\end{itemize}
In all of the datasets except Amazon MR, we follow the data preprocessing
and partitioning in \cite{DBLP:journals/corr/abs-1907-11932}.

\begin{table}[h]
\caption{\label{tab:w3_Explain2Attack_datasets}Statistic of Used Datasets}

\medskip{}

\centering{}{\small{}}%
\begin{tabular}{cccc}
\textbf{Dataset} & \textbf{Train} & \textbf{Test} & \textbf{Avg. Length}\tabularnewline
\hline &  &  & \tabularnewline
IMDB & 25K & 25K & 215\tabularnewline
MR & 9K & 1K & 20\tabularnewline
Amazon MR & 25K & 25K & 100\tabularnewline
Yelp & 560K & 38K & 152\tabularnewline
\end{tabular}{\small{}}{\small\par}
\end{table}

\end{document}